\documentclass[10pt,letterpaper]{article}
\usepackage[top=0.85in,left=2.75in,footskip=0.75in]{geometry}

\usepackage{caption}
\usepackage{subcaption}

\usepackage{amsmath,amssymb}

\usepackage{changepage}

\usepackage[utf8x]{inputenc}

\usepackage{textcomp,marvosym}

\usepackage{cite}

\usepackage{nameref,hyperref}

\usepackage{microtype}
\DisableLigatures[f]{encoding = *, family = * }

\usepackage[table]{xcolor}

\usepackage{array}

\newcolumntype{+}{!{\vrule width 2pt}}

\newlength\savedwidth

\raggedright
\usepackage[aboveskip=1pt,labelfont=bf,labelsep=period,justification=raggedright,singlelinecheck=off]{caption}

\makeatletter
\renewcommand{\@biblabel}[1]{\quad#1.}
\makeatother

\usepackage{lastpage,fancyhdr,graphicx}
\usepackage{epstopdf}
\fancyheadoffset[L]{2.25in}
\fancyfootoffset[L]{2.25in}
\newcommand{\revision}[1]{#1}
\usepackage{lscape} 

\usepackage{appendix}

\begin{document}
\vspace*{0.2in}

\begin{flushleft}
{\Large
\textbf\newline{MGP-AttTCN: An Interpretable Machine Learning Model for the Prediction of Sepsis} }
\newline
\\
Margherita Rosnati\textsuperscript{1},
Vincent Fortuin\textsuperscript{2*}
\\
\bigskip
\textbf{1} Department of Computing, Imperial College London, London, United Kingdom
\\
\textbf{2} Department of Computer Science, ETH Zürich, Zürich, Switzerland
\\
\bigskip

* Corresponding author: \texttt{fortuin@inf.ethz.ch}

\end{flushleft}
\section*{Abstract}
With a mortality rate of 5.4 million lives worldwide every year and a healthcare cost of more than 16 billion dollars in the USA alone, sepsis is one of the leading causes of hospital mortality and an increasing concern in the ageing western world. Recently, medical and technological advances have helped re-define the illness criteria of this disease, which is otherwise poorly understood by the medical society. Together with the rise of widely accessible Electronic Health Records, the advances in data mining and complex nonlinear algorithms are a promising avenue for the early detection of sepsis. This work contributes to the research effort in the field of automated sepsis detection with an open-access labelling of the medical MIMIC-III data set. Moreover, we propose MGP-AttTCN: a joint multitask Gaussian Process and attention-based deep learning model to early predict the occurrence of sepsis in an interpretable manner. We show that our model outperforms the current state-of-the-art and present evidence that different labelling heuristics lead to discrepancies in task difficulty.
\revision{For instance, when predicting sepsis five hours prior to onset on our new realistic labels, our proposed model achieves an area under the ROC curve of 0.660 and an area under the PR curve of 0.483, whereas the (less interpretable) previous state-of-the-art model (MGP-TCN) achieves 0.635 AUROC and 0.460 AUPR and the popular commercial InSight model achieves 0.490 AUROC and 0.359 AUPR.}


\section*{Introduction}

Every year, it is estimated that 31.5 million people worldwide contract sepsis. With a mortality rate of 17\% in its benign state and 26\% for its severe state \cite{fleischmann2016assessment}, sepsis is one of the leading causes of hospital mortality \cite{vincent2014assessment}, costing the healthcare system more than 16 billion dollars in the USA alone \cite{angus2001epidemiology}. Studies demonstrated that early treatment has a significant positive effect on the survival rate \cite{kumar2006duration, nguyen2007implementation}. In particular, \cite{castellanos2010impact} demonstrated that each hour delay in treating a patient results in a 7.6\% increase in mortality. 

Current methods of screening, such as the Modified Early Warning System (MEWS) and the Systemic Inflammatory Response Syndrome (SIRS) have been criticised for their lack of specificity, leading to low accuracies and high false alarm rates. In 2015, the Third International Consensus Definitions for Sepsis \cite{singer2016third, seymour2016assessment, shankar2016developing} committee worked towards incorporating medical and technological advances into an up-to-date definition of sepsis, providing scientists with widely acknowledged illness criteria. Together with the rise of Electronic Health Records (EHR), the scientific community is now armed with both the data and labelling techniques to experiment with novel prediction methods \cite{islam2019prediction, henry2015targeted, ghosh2017septic, calvert2016computational,desautels2016prediction}, which are already proving effective in increasing survival rate \cite{shimabukuro2017effect} and promising in decreasing costs.

The models developed so far either relied on some interpretable yet simple prediction methods, such as logistic regression \cite{calvert2016computational} and decision tree classifiers \cite{mao2018multicentre, delahanty2019development}, or on effective yet black-box methods such as Recurrent Neural Networks \cite{futoma2017improved}. Moreover, the results achieved by different authors are rarely comparable: although most use the MIMIC-III data set, the disparities in labelling rules result in highly variable data sets (eg. \cite{raghu2018model} have 17,898 septic patients vs.\ 2,577 for \cite{desautels2016prediction}).  

This work presents an attempt at reconciling interpretability and predictive performance on the sepsis prediction task and makes the following contributions:

\begin{itemize}
\item Gold standard for labelling. We provide a gold standard for Sepsis-3 labelling implemented on the MIMIC-III data set.
\item Novel interpretable model. We present an explainable and end-to-end trainable model based on Multitask Gaussian Processes and Attentive Neural Networks for the early prediction of sepsis.
\item Empirical evaluation. We assess our model on real-world medical data and report superior predictive performance and interpretability compared to previous methods.
\end{itemize}

\begin{figure*}[!h]
    \centering
    \includegraphics[width=\linewidth]{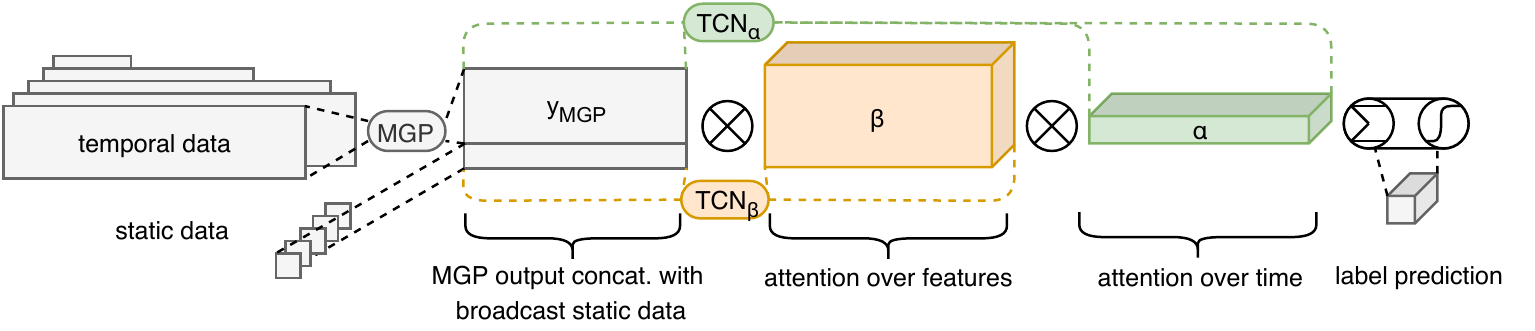}
\caption{\bf Proposed model architecture.}
    \label{fig:model}
\end{figure*}

An overview of our proposed method is shown in Figure~\ref{fig:model}.
The code for labeling the data\footnote{\url{https://github.com/mmr12/MIMIC-III-sepsis-3-labels}} and for running the models\footnote{\url{https://github.com/mmr12/MGP-AttTCN}} is publicly available.

\section*{Related work}

\paragraph{Medical time series diagnostics} Multiple researchers have tackled the task of predicting sepsis and septic shock. Works on septic shock include exploration of survival models \cite{henry2015targeted} and Hidden Markov Models \cite{ghosh2017septic}. However, these models rely on the assumption that a patient has already developed sepsis and focus on patterns of patients' further deterioration. Other authors  \cite{calvert2016computational, desautels2016prediction,mao2018multicentre, delahanty2019development} use linear models and decision trees on engineered features to predict sepsis onset, thus failing to capture temporal patterns. More recently, \cite{kam2017learning} and \cite{raghu2018model} employed recurrent neural networks to better capture time dependencies. Crucially, all these models rely on either averaging or forward imputation of data points to create equidistant inputs. This transformation creates data artefacts and discards relevant uncertainty: in the medical field, the absence of data is a conscious decision made by professionals implying an underlying belief of the patient state. \cite{futoma2017improved} and \cite{moor2019early} tackled this issue with Multitask Gaussian Processes (MGPs), however their models lack the interpretability necessary in the medical field.

\paragraph{Irregularly sampled time series} The most common solution to missing values is forward imputation \cite{calvert2016computational}. \cite{lipton2016modeling} utilise forward imputation coupled with a missingness indicator fed into a  black-box model. Although this method retains information about data presence, it is not clear how the information is later interpreted by the model and hence does not meet our transparency criteria. \cite{ghassemi2015multivariate} use MGPs to fit sparse medical data, however they optimise their model for the data fit and use the parametrisation as input for a classifier rather than optimising the model for a classification task. Both \cite{futoma2017learning} and \cite{moor2019early} use MGPs with end-to-end training, although their temporal covariance function is shared across all variables. Finally, \cite{futoma2017improved} uses MGPs with multiple time kernels in a similar fashion to our model, although they infer the number of kernels from hyperparameter tuning rather than the data itself.

\paragraph{Attention based neural networks} Attention was first introduced on the topic of machine translation \cite{bahdanau2014neural}. Since then, the concept has been used in natural language processing \cite{yang2016hierarchical, yu2018mattnet} and image analysis \cite{mnih2014recurrent, schlemper2019attention}. In the same spirit, \cite{qin2017dual} used attention mechanisms to improve the performance of a time series prediction model. Although their model easily explains the variable importance, its attention mechanism is based on Long Short Term Memory encodings of the time series. At any given time, such an encoding contains both the information of the current time point and all previous time points seen by the recurrent model. As such, the time domain attention does not allow for easy interpretation. More similar to our implementation is the RETAIN model \cite{choi2016retain}, which generates its attention weights through reversed recurrent networks and applies them to a simple embedding of the time series. The model employs recurrent neural networks which are slower to train and suffer from the vanishing gradient problem. Furthermore, the initial and final embeddings decrease the model's interpretablity. Attention in combination with a Temporal Convolutional Network (TCN) has also been used by \cite{lin2019medical}, but there it only attends to time points and not to different features.

\revision{
\paragraph{Parallel work on sepsis prediction}
As mentioned above, sepsis prediction on the ICU is an important and timely problem and an active area of research. Under these circumstances, it is not surprising that some approaches have been developed in parallel to this work \cite{kong2020using, hou2020predicting, acsurouglu2021deep, yao2020machine, kok2020automated, li2020real, song2020predictive, svenson2020sepsis, lauritsen2020early, narayanaswamy2019machine, chaudhary2021outcome}.
It will be an exciting and important avenue for future work to benchmark all these approaches (including ours) against each other and to compare their performances on a unified and realistic set of sepsis labels, for instance, the ones we propose in this work.
}

\section*{Method}

In the following, we will provide a detailed explanation of our proposed model and its different components.
A graphical overview of the model is shown in Figure~\ref{fig:model}.

\subsubsection*{Notation}
Let us first define some notation for the problem at hand.
For each patient encounter $p$, several features $y_{p,t_i, k}$ are recorded at times $t_{p,k,i}$ from admission, where $k\in\{1,\dots ,M\}$ is the feature identifier. These features are often vital signs and laboratory results. As such, they are rarely observed at the same times. Hence, we have a sparse matrix representation of observations
\begin{equation}
    \begin{pmatrix} 
y_{p,1,t_1} & \dots & y_{p,1,t_{N_p}}\\
\vdots &\ddots & \\
y_{p,M,t_1} & \dots & y_{p,M,t_{N_p}}
\end{pmatrix}
\end{equation}
where $N_p$ is the patient's observation period length.
We also define static features $\mathbf{s}_p=\{s_{p, M+1}, .. s_{p, M+Q}\}$ with feature identifiers $k\in\{M+1,\dots ,M+Q\}$, corresponding to time-independent quantities, such as age, gender and first admission unit.
Finally, we define sepsis labels $l_p\in\{0,1\}$. Given the sparsity of the data, we can define the compact representation of all observed values: 
\begin{equation}
 \{\mathbf{t}_p, \mathbf{y}_p, \mathbf{s}_p, l_p\} 
    =  \left\{
    \begin{array}{l}
    \{t_{p,i,k}\}_{i \in \{0,\dots,N_p\}, k \in \{1,\dots,M\}},\\
    \{y_{p,i,k}\}_{i \in \{0,\dots,N_p\}, k \in \{1,\dots,M\}},\\
    \quad \{s_{p, M+1}, .. , s_{p, M+Q}\},\\
    \quad l_p
    \end{array}
\right\}
\end{equation}

The goal of the model is, for a given set $\{\mathbf{t}_p, \mathbf{y}_p, \mathbf{s}_p\} $ to predict the label $l_p$. In order to remove clutter, we will from now on drop the patient-specific subscript $p$ from all notation, and the feature subscript $k$ from time notation, simplifying $t_{p,k,i}$ to $t_i$.

\subsubsection*{Multitask Gaussian Process (MGP)}
\label{sec:methodology:MGP}

Gaussian processes are non-parametric Bayesian models commonly known for their ability to generate coherent function fits to a set of irregular samples, by modelling the data covariance. As they easily account for uncertainty and do not require homogeneously sampled data, Gaussian processes are the perfect candidate model to deal with irregularly sampled medical time series. 

We use a Multitask Gaussian Process (MGP) \cite{bonilla2008multi} to capture feature correlation and \cite{li2016scalable}'s end-to-end training framework, in a similar manner to \cite{futoma2017learning}. Given an hourly spaced time series $\{t_{i}'\}_{i=-N_p}^0$ where 0 is the time of prediction, the MGP layer produces a set of posterior predictions for each feature, which will then be fed into a classification model.

We define a patient-independent prior over the true values of $\{y_{i,k}\}$ by $\{f_{k}(t_{i})\}$ such that 
\begin{gather}
    y_{i,k} 
\sim \mathcal{N}
(f_{k}(t_i) ,   
\sigma_{k}^2)\\
    \big<   f_{k}(t_{i}), f_{k'}(t_{j}) \big> =
    \sum_{l\in L} K_l^\mathbf{k}(k, k')\,K_l^{\mathbf{tt}}(t_{i}, t_{j})
\end{gather}
where $\{K_l^{\mathbf{tt}}(t_{i}, t_{j})\}_{l \in L}$ are parametric time point covariances varying in smoothness, $\{K_l^\mathbf{k}(k, k')\}_{l \in L}$ are free-form feature covariances at a given smoothness level, independent of time, and $L$ are smoothness clusters. 
Over all variables and time points, the multivariate model has covariance
\begin{equation}
 \sum_{l\in L}K_l^\mathbf{k} \otimes K_l^{\mathbf{tt}} + D \otimes I   
\end{equation}
where $D = diag(\sigma_{k})$ are the noise terms associated to each feature and $\otimes$ is the Kronecker product. This formulation allows each datapoint to be defined as both a function of its own timed observations and observations of the remaining features. 

The quick illness progression is well suited to the quadratic growth of the covariance matrix. However, for the few cases when a patient develops sepsis well into their hospitalisation, a suitable measure to prevent excessive memory consumption is to ignore the initial datapoints of the patient. 

Note that there are two main feature clusters: vital signs (vitals) and laboratory results (labs). Vitals are noisier and sampled more often, whereas labs are more monotone and rarely sampled. 
As opposed to \cite{futoma2017improved}, we do not treat the number of clusters $L$ as hyperparameters but set $L=2$ and define
\begin{align}
 K_l^t(t_{i}, t_{j}) = \text{exp}\Big(\frac{-|t_{i} - t_{j}|}{\lambda_l}\Big)   
\end{align}
as Ornstein-Uhlenbeck (OU) kernels with lengths $\lambda_1$ and $\lambda_2$, each representing a cluster smoothness. OU kernels are well suited to capture local variations and do not assume infinite differentiability as Squared Exponential kernels do. In our case, differentiablity implies a level of smoothness which is unrealistic for medical records and only introduces unnecessary bias. In addition, given the scarce availability of labs, short lengthscales would be an ill fit to the data. We hence discarded kernels varying over lengthscales such as the Cauchy and the Rational Quadratic kernels. $K_l^k(k, k')$ are free-form covariance matrices that are learned by gradient descent.

The posterior over the reguarly sampled timepoints $\mathbf{t}' = \{t_{i}'\}_{i=-N_p}^0$ is a multivariate Gaussian with mean and covariance:
\begin{equation}
    \begin{split}
        \boldsymbol{\mu} =& \big(\sum_{l\in L} K_l^\mathbf{k} \otimes K_l^{\mathbf{tt'}}\big)\big(\sum_{l\in L}K_l^\mathbf{k} \otimes K_l^{\mathbf{tt}}  + D \otimes I\big)^{-1}\mathbf{y}\\
        \Sigma =& \sum_{l\in L}K_l^\mathbf{k}\otimes K_l^{\mathbf{t't'}}\\
        &- \big(\sum_{l\in L} K_l^\mathbf{k} \otimes K_l^{\mathbf{tt'}}\big)\big(\sum_{l\in L}K_l^\mathbf{k} \otimes K_l^{\mathbf{tt}}  + D \otimes I\big)^{-1}\big(\sum_{l\in L} K_l^\mathbf{k} \otimes K_l^{\mathbf{t't}}\big) 
    \end{split}
\end{equation}
In order to approximate the posterior distribution, we then take Monte Carlo samples $\mathbf{y}_{\text{MC}}$ from $\mathbf{Y}_{\text{MGP}} \sim \mathcal{N}(\boldsymbol{\mu}, \Sigma)$.

To feed the MGP samples into the classifier, we fix the model time window to $N=25$ by either zero padding or truncating the beginning of the time series. We choose to do so at the beginning of the time series in order to align prediction times as the last step of the temporal classification model.
Here, we also integrate the static variables by broadcasting them over each time. The reasoning behind this design choice is explained in more details in the following section.

\subsubsection*{Attention Time Convolutional Network (AttTCN)}
\label{sec:att-tcn}

The concept of attention was born in machine translation \cite{bahdanau2014neural} and has recently successfully been applied to different types of sequential data \cite{yang2016hierarchical, yu2018mattnet,mnih2014recurrent, schlemper2019attention,qin2017dual}. 
In machine translation, given an input sentence embedding 
\begin{align}
  S = \{ \mathbf{h}_1, \dots, \mathbf{h}_{|S|}\}  
\end{align}
the attention mechanism produces weights 
\begin{align}
  \{\alpha^i_1, \dots, \alpha^i_{|S|}\}\quad\quad\text{  s.t.  }\quad\quad \alpha^i_j \in [0,1]\, , \quad\sum_j\alpha^i_j=1  
\end{align}
and a context vector 
\begin{align}
  \mathbf{c}_i = \sum_{j}\alpha_j^i\mathbf{h}_j  
\end{align}
used to predict target word $i$. The weights $\alpha_j^i$ can therefore be interpreted as the importance of the input sentence's $j^{\text{th}}$ word to produce the $i^{\text{th}}$ word of the translation.

More recently, \cite{choi2016retain} have applied attention to clinical time series. Given a time series 
\begin{align}
 \{\mathbf{x}_1,\dots, \mathbf{x}_T\} \subset \mathbb{R}^r \; ,
\end{align}
the authors first create a time-independent embedding of the data 
\begin{align}
 \{\mathbf{v}_1, \dots, \mathbf{v}_T\} \subset \mathbb{R}^m  \; .
\end{align}
They then use inversed recurrent neural networks (RNN) to create weights $\boldsymbol{\alpha} \in \mathbb{R}^T$ and $\boldsymbol{\beta}\in\mathbb{R}^{T\times m}$, where $\alpha_j\in[0,1]$ and $\beta_{ij}\in[-1,1]$, with softmax and tanh activations respectively. The context vectors then take the form 
\begin{align}
 c_i = \sum_{j\leq i}\alpha_j\boldsymbol{\beta}_j\odot v_j \; ,  
\end{align}
where $\odot$ is the element-wise product, and are fed into a multilayer perceptron with softmax activation to yield a prediction.

The attention model we devised borrows some ideas from \cite{choi2016retain}. Two embeddings, $\mathbf{z} = [z_1,\dots,z_N]$ and $\mathbf{z'} = [z'_1,\dots,z'_N]$ with $\mathbf{z},\mathbf{z'} \in \mathbb{R}^{N\times (M+Q)}$, are directly generated from the interpolated data $\mathbf{y}_{\text{MC}} \in \mathbb{R}^{N\times (M+Q)}$ through two temporal convolutional networks (TCNs), removing a layer of abstraction and hence facilitating interpretability.

TCNs are a class of neural networks composed of \textit{causal} convolutions stacked into Residual Blocks. A causal convolution is a 1D convolutional layer which only takes inputs from the past to generate its output, avoiding any information leakage from the future. Residual Blocks are made of two causal convolutional layers together with ReLU activation functions, dropout and L2 regularisations. The Residual Blocks also include an identity map from the input of the block added to the output. As we only use up to 12 layers, this last step is omitted in our architecture.
TCNs have shown to outperform RNNs \cite{bai2018empirical}, are faster at training and do not suffer from vanishing gradients. Given the latter, inverting the time series similarly to \cite{lea2017temporal} also becomes an unnecessary step which we omit.

We generate the attention weights $\mathbf{\alpha}$ and $\mathbf{\beta}$ as
\begin{align}
    &\alpha_{j,0} =\text{softmax}( \mathbf{z}_j \times \mathbf{W}_{\alpha, 0} + b_{\alpha,0})\\
    &\alpha_{j,1} =\text{softmax}( \mathbf{z}_j \times \mathbf{W}_{\alpha, 1} + b_{\alpha,1})\\
    &\boldsymbol{\beta}_{j,0} = \text{sigmoid}(\mathbf{z}_j' \times \mathbf{W}_{\beta, 0} + \mathbf{b}_{\beta,0})\\
    & \boldsymbol{\beta}_{j,1} = \text{sigmoid}(\mathbf{z}_j' \times \mathbf{W}_{\beta, 1} + \mathbf{b}_{\beta,1})
    \end{align}\begin{align}
    & \mathbf{W}_{\alpha, 0},\mathbf{W}_{\alpha, 1}\in\mathbb{R}^{M+Q} & &  b_{\alpha,0},  b_{\alpha,1}\in \mathbb{R}\\
    & \mathbf{W}_{\beta, 0},\mathbf{W}_{\beta, 1} \in \mathbb{R}^{(M+Q)\times (M+Q)}
    & & \mathbf{b}_{\beta,0},  \mathbf{b}_{\beta,0}\in\mathbb{R}^{M+Q} \; ,
\end{align}
such that $\boldsymbol{\alpha} = [\boldsymbol{\alpha}_0,\boldsymbol{\alpha}_1]\in\mathbb{R}^{N\times2}$ and $\boldsymbol{\beta}=[\boldsymbol{\beta}_0,\boldsymbol{\beta}_1]\in\mathbb{R}^{N\times (M + Q) \times2}$. 

We then create two context vectors, one for each of the negative and positive label predictions 
\begin{equation}
    \mathbf{c}_i=\sum_{j\leq i}\alpha_{j,\delta}\boldsymbol{\beta}_{j,\delta} \odot \mathbf{y}_{\text{MC},j}\in\mathbb{R}^{N\times(M+Q)\times 2} \; ,\quad \delta\in\{0,1\} \; ,
\end{equation}
where $\mathbf{y}_{\text{MC},j}$ is broadcast to meet the dimensionality of $\boldsymbol{\beta}_{j,\delta}$. We then predict the labels as
\begin{equation}
 \hat{\mathbf{l}}_i = \text{softmax}\Big(\sum_{n}^{N}\sum_m^{M+Q}\mathbf{c}_{i,nm}\Big) \in [0,1]^2 \; .
\end{equation}

In our case, we are only interested in making predictions with the latest available data. We therefore only use $\mathbf{\hat{l}}_{\text{last}}$ to train the model. This of course can be easily modified to suit any specific use case.

Since the MGP output is directly multiplied by weights $c_i$, the classification model can be interpreted as a scoring mechanism where each past point $y_{\text{MC},ij}$ contributes $\alpha_{i,0}\beta_{ij,0}$ to the time series being classified as positive, and $\alpha_{i,1}\beta_{ij,1}$ to the time series being classified as negative. The positive and negative scores are then normalised to represent probabilities of the positive or negative labelling. As we designed both $\boldsymbol{\alpha}$ and $\boldsymbol{\beta}$ to be non-negative, we can hence directly look at the average $\boldsymbol{\alpha}$ and $\boldsymbol{\beta}$ over Monte Carlo samples to see which time points and features contribute most strongly to the network's positive or negative decision.
This facilitates the interpretability of our model compared to previous approaches.

\section*{Data}

Sepsis is defined as a life-threatening organ dysfunction caused by a dysregulated host response to infection \cite{singer2016third}. A dysregulated host response is interpreted as a suspicion of infection. In EHR terms, it is encoded by the administration of high spectrum antibiotics and a bacterial blood culture within a set interval of each other. The organ dysfunction is interpreted as a two point increase in Sequential Organ Failure Assessment (SOFA) within a suspected infection window. The SOFA score quantifies the deterioration of different systems - respiration, coagulation, liver function, cardiovascular function, central nervous system, and renal function. 

We make use of the MIMIC-III dataset, a collection of medical records for over 40'000 patients who stayed in critical care units of the Beth Israel Deaconess Medical Center between 2001 and 2012 \cite{johnson2016mimic}. The records are composed of vital sign recordings, laboratory tests, drugs administered, and patients' outputs. We encode the Sepsis-3 criteria in the MIMIC-III dataset following \cite{johnson2016mimic,alistair_johnson_2018_1256723}'s code available on GitHub and \cite{moor2019early}'s code that the authors have generously provided.

One key difference between our assumptions and the ones \cite{moor2019early} develop is the handling of missing SOFA contributor values: if one or more SOFA contributors are missing, \cite{moor2019early} do not calculate the total score. On the other hand, we assume such a contributor to be within a healthy norm, hence implying a zero contribution. With our methodology, patients worsening in one area but with no measurements in another will be considered septic, whereas they will not in the \cite{moor2019early} dataset.
\revision{After discussions with a clinician, we learned that the standard practice in the clinic would be to assume healthy values for all unmeasured variables, as we did in our labeling.
Moreover, if the treating physician would expect a variable to be outside the healthy norm, they would usually measure it, such that most unmeasured variables will indeed have a high probability of being healthy.
Our labeling approach thus fits better with the standard clinical practice and includes more septic patients that would not be included in \cite{moor2019early}'s data.
Moreover, we hypothesize that these patients would be the ones where the treating clinicians do not already suspect a sepsis and have thus not measured all the SOFA variables, which will potentially make them harder to classify but arguably also more interesting, because they would otherwise be missed by the treating doctors.}

In order to validate our results, we carry out all experiments on using both labelling techniques.

\paragraph*{Patient Inclusion}
We filter for patients admitted to Intensive Care Units (ICU) who are more than 14 years old and with valid records. Case patients are patients having sepsis onset within their ICU stay, whereas control patients have not developed sepsis nor have an ICD discharge code referring to sepsis. Starting with \revision{58,976} patients, we find \revision{14,071} control patients and \revision{7,936} case patients using our labels, versus \revision{1,797} cases using \cite{moor2019early}'s labels.

\paragraph*{Feature extraction}
Reviewing sepsis-related literature and commonly extracted laboratory and vital recordings, we extract all features which were reported at least once for more than 75\% of the included population. The final 24 dynamic features are reported in Table \ref{tab:dym_feat}.
We also extract static features -- age, gender, and first ICU admission department.

\begin{table}[!ht]
\caption{List of dynamic features}
\begin{tabular}{|c|c c|}
\textbf{Vitals} & \multicolumn{2}{c|}{\textbf{Labs}} \\ \hline
\revision{Systolic blood pressure (sysbp)} & bicarbonate & \revision{Pulse transit time (ptt)}\\
\revision{Diastolic blood pressure (diabp)} & creatinine & \revision{International normalized ratio (inr)}\\
\revision{Mean blood pressure (meanbp)} & chloride & \revision{Prothrombin time (pt)}\\
\revision{Respiratory rate (resprate)} & glucose & sodium\\
\revision{Heart rate (heartrate)} & hematocrit & \revision{Blood urea nitrogen (bun)}\\
\revision{Pulse oximetry (spo2 pulsoxy)} & hemoglobin & \revision{White blood cells (wbc)}\\
\revision{Body temperature (tempc)} & lactate & magnesium\\
& platelet & \revision{Blood gas pH (ph bloodgas)}\\
& potassium &\\

\end{tabular}
\label{tab:dym_feat}
\end{table}

\paragraph*{Case-control matching}
As the goal is to predict sepsis prior to onset, the cases data was extracted between ICU admission and sepsis onset. Note that sepsis onset happens early within ICU admission, with the median patient getting sick at 3.4 hours after admission. On the other hand, patients not developing sepsis are more likely to recover completely, and do so in a lengthier time frame. In addition, once they are close to discharge, their vitals and labs are within the norms. Hence, both the length and the values of the time series are strong discriminatory factors which ease the classification. \revision{We hence carry out a matching strategy similar to \cite{moor2019early}: following the class imbalance ratio, we associate each control time series to a case time series and truncate the control to have the same length as the case from ICU admission. We then discard patients with less than 40 data points within the selected window, and---for computational tractability---truncate the first $N_p - 250$ initial values of patients' time series in order to keep a maximum of 250 data points per patient.}

\begin{figure*}[!h]
    \centering
    \includegraphics[width=\linewidth]{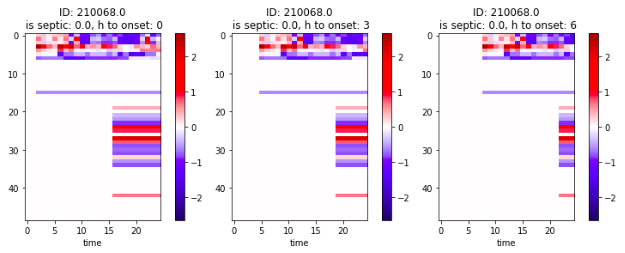}
\caption{{\bf Baseline patient data for different horizons.} x-axis: Time from admission. y-axis: Feature identifier.}
    \label{fig:forward_imp}
\end{figure*}

\paragraph*{Horizon augmentation}
As our goal is  to predict sepsis early, we augment the data by creating new shorter time series. For each time series, we create six copies, where each copy represents a different horizon to onset. We then proceed to truncate the last one to six hours prior to onset from the time series copies. In order to keep data consistency, we once again discard time series with less than 40 observations. Figure \ref{fig:forward_imp} is a graphical representation of the discretised version used for the baseline of an augmented datapoint, whereas in Tables \ref{tab:all_pats_Moor} and \ref{tab:all_pats_Rosnati} we illustrate the data distribution per horizon. 

\begin{table}[!ht]
\caption{\bf Augmented dataset statistics with Moor et al. \cite{moor2019early} labels.}
\begin{tabular}{c|cc}

{Horizon to onset} & N. of patients & N. of obs. per patient \\ \hline
0 h & \revision{15,123} &$ 69.9 \pm 59.6 $\\ 
1 h & \revision{11,258} &$ 56.6 \pm 59.1 $\\ 
2 h & \revision{8,478} &$ 61.4 \pm 62.8 $\\ 
3 h & \revision{6,554} & $ 66.5 \pm 65.9 $\\ 
4 h & \revision{5,233} & $ 70.6 \pm 69.0 $\\ 
5 h & \revision{4,162} & $ 76.3 \pm 71.9 $\\ 
6 h & \revision{3,390} & $ 81.9 \pm 74.3 $\\ 
\end{tabular}

\label{tab:all_pats_Moor}
\end{table}

\begin{table}[!ht]
\caption{\bf Augmented dataset statistics with our labels.}
\begin{tabular}{c|cc}

{Horizon to onset} & N. of patients & N. of obs. per patient \\ \hline
0 h & \revision{20,075} &64.0 $\pm$ 65.5 \\ 
1 h & \revision{15,832} &62.5 $\pm$ 65.6 \\ 
2 h & \revision{12,080} &66.1 $\pm$ 67.2 \\ 
3 h & \revision{9,441} &69.7 $\pm$ 68.1 \\ 
4 h & \revision{7,484} & 73.4 $\pm$ 68.5 \\ 
5 h & \revision{6,007} &77.1 $\pm$ 68.2 \\ 
6 h & \revision{4,876} &81.2 $\pm$ 67.4 \\ 
\end{tabular}

\label{tab:all_pats_Rosnati}
\end{table}

\paragraph*{Data split}
Finally, we split the data into training, validation, and testing sets, respectively capturing 80\%, 10\%, and 10\% of the data. We then normalise the data by subtracting the training set mean and dividing by the training set standard deviation of each feature.

\paragraph*{Hyperparameter search and training}
As the datasets are highly imbalanced, we carry out a case set oversampling: we randomly resample the case set to have the same size as the control set. In addition, at each iteration we sample the same number of cases and controls and then feed a shuffled version into the model. In this manner, the model will see an equal number of controls and cases and will not become biased towards zero labels. This procedure is not applied to the validation and test sets, as the results would not compare to real-life settings.

For both our model MGP-AttTCN and all baselines, in order to select the best possible hyperparameters, we performed a hyperparameter random search, as described in Table \ref{tab:hyperparam}.

\begin{table}[!ht]
\caption{\bf Hyperparameter search.}
\begin{tabular}{l|c|c}
\multicolumn{1}{c|}{Hyperparameter} & \multicolumn{2}{c}{Random Search} \\
 & min & max \\\hline
MGP Monte Carlo samples & 4 & 20 \\
TCN kernel size & 2 & 6 \\
TCN number of Residual Blocks & 2 & 12 \\
TCN number of hidden layers & 10 & 55 \\
TCN dropout rate & 0 & 0.99 \\
TCN L2 regularisation & 0 & 250
\end{tabular}
\label{tab:hyperparam}
\end{table}

\section*{\revision{Baselines}}
\label{appendix:baselines}
\subsection*{\revision{Data preparation}}
\label{appendix:bl:data_prep}

\revision{
In order to benchmark our MGP model, we build some baselines homogenising the data sampling. For each hour and variable, we take the average of the available observations. If a given hour has no observations, we carry forward the average of the previous hour. In this manner, we generate an hourly sampled time series for each patient. We then proceed to normalise the size of each patient matrix by setting a time window of observation $N$. For patients having more than $N$ observations $N_p$, we discard the first $N-N_p$ observation; whereas for patients having less than $N$ observations, we pad the beginning of the matrix with zeros.}

\revision{
\begin{align}
\begin{pmatrix} 
y_{p,1,t_1} & \dots & y_{p,1,t_{N_p}}\\
\vdots &\ddots & \\
y_{p,M,t_1} & \dots & y_{p,M,t_{N_p}}
\end{pmatrix}
\xrightarrow{\text{carry forward}}&
    \begin{pmatrix} 
y_{p,1,1} & \dots & y_{p,1,N_p}\\
\vdots &\ddots & \\
y_{p,M,1} & \dots & y_{p,M,N_p}
\end{pmatrix}
\end{align}
\begin{align}
\xrightarrow{\text{normalise}}&
\begin{cases}
   & \begin{pmatrix} 
y_{p,1,N-N_p} & \dots & y_{p,1,N}\\
\vdots &\ddots & \\
y_{p,M,N-N_p} & \dots & y_{p,M,N}
\end{pmatrix}
\text{   if }N_p \geq N\\
   & \begin{pmatrix} 
0 & \dots &0 & y_{p,1,N-N_p} & \dots & y_{p,1,N}\\
\vdots & & \vdots & \vdots &\ddots & &\\
0 & \dots &0 & y_{p,M,N-N_p} & \dots & y_{p,M,N}
\end{pmatrix}
\text{   else}\\
\end{cases}
\end{align}
}

\revision{
We choose to align the end of the time series as opposed to the beginning, as the relative importance of time points is higher closer to when a patient becomes sick rather to when they are admitted to the ICU.
}

\revision{
As a next step, we augment the data to focus on different time series in a similar manner as for irregularly sampled data. We create seven copies of each time series, where for each copy we discard the last zero to six hours, then normalise the matrix as above. We hence generate a dataset 
$\mathbf{Y}_{\text{BL}} = \{Y\}_q= \{\{y_{q,ij}\}_{i,j=1}^{N,M} \}_q$ where $q$ represents all augmented the time series.
}
\subsection*{\revision{InSight}}
\label{appendix:bl:InSight}

\revision{
The \emph{InSight} scoring model is one of the few machine learning algorithms to surpass the proof-of-concept stage with multiple research, economic and clinical trials \cite{calvert2016computational, desautels2016prediction, calvert2017cost, mao2018multicentre}. We therefore include it as a baseline to our model.
The key concept of the model is to use few largely available vitals, build some handcrafted features and train a simple classification model.
}

\revision{
Here we provide a brief summary of the main method.
The features extracted are based on a window of six consecutive hours. For each six hour window, we extract each variable's mean $M_i$ and difference $D_i$ (last observation minus first observation) over the window. We also extract variable pairs correlation $D_{ij}$ and \textit{triplet correlation} $D_{ijk}$, where $i, j, k$ are observed variables. We interpret the latter as a relaxation of the Pearson correlation: if the correlation between two variables is 
\begin{equation}
    \rho_{XY} = \frac{\mathbb{E}[(X - \mu_X)(Y - \mu_Y)]}{\sigma_X\sigma_Y}
\end{equation}
then we define the \textit{triplet correlation} as 
\begin{equation}
    \rho_{XYZ} = \frac{\mathbb{E}[(X - \mu_X)(Y - \mu_Y)(Z - \mu_Z)]}{\sigma_X\sigma_Y\sigma_Z}
\end{equation}
We then classify the difference and correlations as either positive, negligible or negative using their distribution quantiles over every patient and six hour window observed. Note that given the high level of data missingness, many variables are calculated by forward imputation and hence have no variance over six hours. To adjust for the high number of zero correlations, we calcualte the quantiles of non-zero correlations and define:
\begin{equation}
    \hat{D}_{i} = 
    \begin{cases}
    1 \quad &\text{if } D_i > q^*(2/3) \\
    -1 \quad &\text{if } D_i < q^*(1/3) \\
    0 \quad &\text{otherwise}
    \end{cases}
\end{equation}
where $q^*$ is the adjusted quantile function. We proceed in a similar manner for the correlations and triplet correlations.
}

\revision{
In order to keep the results comparable to the AttTCN fixed window $N$, we extract $N - (6-1)$ six consecutive hour window and vectorise the resulting features, generating in total
\begin{equation}
    n_{\text{features}} = \Big(N - 5\Big)\times \Big(2\times M + {M \choose 2} + {M \choose 3}\Big)
\end{equation}
features per patient.
}

\revision{To remain consistent with the original work, we only kept patients with at least one observation for each feature over the 5 hour period for the following observations: age, systolic blood pressure, pulse transit time, heart rate, temperature, respiration rate, white blood cell count, pH and pulse oximetry. The corresponding dataset statistics can be found in Tables \ref{tab:all_pats_Rosnati_baselines} and \ref{tab:all_pats_Moor_baselines}.}

\begin{table}[!ht]
\caption{\bf \revision{Baseline dataset with our labels.}}
\begin{tabular}{c|cc}

\revision{Horizon to onset} & \revision{N. of patients} & \revision{N. of patients with onset} \\ \hline
\revision{0 h} & \revision{1,943} & \revision{786 (40.5\%)} \\
\revision{1 h} & \revision{1,938} & \revision{782 (40.4\%)} \\
\revision{2 h} & \revision{1,810} & \revision{690 (38.1\%)} \\
\revision{3 h} & \revision{1,657} & \revision{606 (36.6\%)} \\
\revision{4 h} & \revision{1,504} & \revision{532 (35.4\%)} \\
\revision{5 h} & \revision{1,352} & \revision{467 (34.5\%)} \\
\revision{6 h} & \revision{1,217} & \revision{420 (34.5\%)} \\
\end{tabular}

\label{tab:all_pats_Rosnati_baselines}
\end{table}

\begin{table}[!ht]
\caption{\bf \revision{Baseline dataset with Moor et al. \cite{moor2019early} labels..}}
\begin{tabular}{c|cc}

\revision{Horizon to onset} & \revision{N. of patients} & \revision{N. of patients with onset} \\ \hline
\revision{0 h} & \revision{2,298} & \revision{658 (28.6\%)} \\
\revision{1 h} & \revision{2,294} & \revision{657 (28.6\%)} \\
\revision{2 h} & \revision{2,187} & \revision{609 (27.8\%)} \\
\revision{3 h} & \revision{1,981} & \revision{531 (26.8\%)} \\
\revision{4 h} & \revision{1,748} & \revision{453 (25.9\%)} \\
\revision{5 h} & \revision{1,518} & \revision{384 (25.3\%)} \\
\revision{6 h} & \revision{1,330} & \revision{329 (24.7\%)} \\
\end{tabular}

\label{tab:all_pats_Moor_baselines}
\end{table}

\revision{
Although the original paper does not specify which classification method the authors employ, we derive by their description of a \textit{dimensionless score} that the method is a logistic regression.
}

\subsection*{\revision{Other baselines}}
\paragraph*{\revision{Logistic Regression}}
\revision{As a simple baseline, we perform a Ridge Logistic Regression using the hourly data described above.}
\paragraph*{\revision{Ablation models}}
\revision{
In addition to Insight and the Logistic Regression, we perform ablation studies on our proposed model. In a first instance, we remove the AttTCN arm and replace it by a logistic regression (model ``MGP-log.reg."). Secondly, we remove the arm of the AttTCN controlling the attention over time $\alpha$ (model ``MGP-AttTCN w/o $\alpha$"), then the arm generating $\beta$ (model ``MGP-AttTCN w/o $\beta$"). }
\paragraph*{\revision{MGP-TCN}}
\revision{Finally, we compare our model to the state-of-the-art MGP-TCN algorithm \cite{moor2019early}.}
\section*{Experimental Results}

\begin{figure*}[!h]
    \centering
	\includegraphics[width=\linewidth]{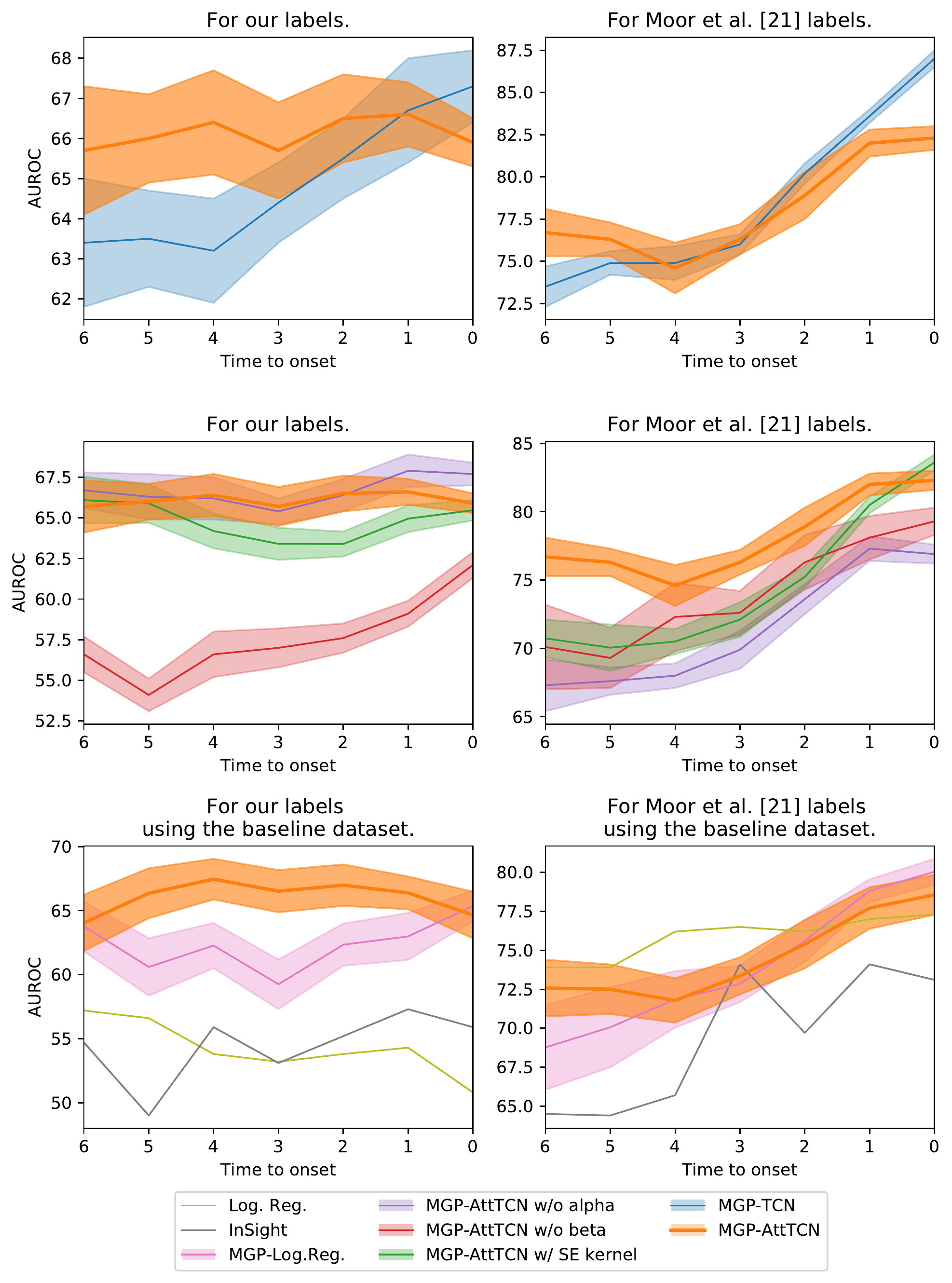}
    \caption{{\bf \revision{Area under the ROC curve }of different models.} It can be seen that our proposed labels are harder to fit than the ones by \cite{moor2019early}. Moreover, our proposed model outperforms the baselines on both label sets, especially for earlier prediction horizons.}
    \label{fig:roc_results}
\end{figure*}

We compare our model's performance to the performance of the InSight algorithm \cite{calvert2016computational} and to the state-of-the-art MGP-TCN algorithm \cite{moor2019early}. 
Figure \ref{fig:roc_results} shows the predictive performance of the models for different time horizons, whereas \revision{the numerical results can be found in the Appendix, in Tables~S1-8.}

\subsubsection*{Comparison between different data labels}

The first result is the difference in performance of models applied to the different labelling methods. The SOFA contributor assumption from \cite{moor2019early} has two main implications. Firstly, it considerably restricts the number of patients. Assuming that sicker patients receive more medical attention, the patients included are likely to be in worse conditions than the septic patients excluded and hence easier to classify. Secondly, it delays sepsis onset. For example, a patient having a severe liver failure with few other recorded vitals, followed by an overall collapse further in time will have septic onset at the time of its liver failure in our records, whereas it will only be considered septic at the time of the overall collapse in \cite{moor2019early}'s labels. On the other hand, the labels we produce reflect the incomplete nature of medical data: even if only a part of all the potentially relevant tests are carried out at any given time, a doctor must be able to assess a patient's well-being and foresee potential complications.
The difference in labels implies a discrepancy in task difficulty: \cite{moor2019early}'s labels present an easier learning problem, but define a more narrow use case in real-world scenarios.

Indeed, when assessing the performance of the different models on the two different data labellings, it becomes evident that our proposed labels are harder to fit.
This means that predicting sepsis in a realistic setting on the intensive care unit is probably much harder than previous work would suggest.

\subsubsection*{Model performance}

We find that our MGP-AttTCN model yields a better performance \revision{than the MGP-TCN \cite{moor2019early}} when presented with patients further in time from sepsis onset (i.e., earlier in time) \revision{(Fig.~\ref{fig:roc_results}, top row)}. \revision{In the case of our labels the difference is clearly noticeable, whereas with \cite{moor2019early}'s labels it is of lower statistical significance.}

\revision{Moreover, we observe that ablations of our model (e.g., changing the GP kernel, removing the $\alpha$ weights, or removing the $\beta$ weights) reduces our model's performance, as expected (Fig.~\ref{fig:roc_results}, middle row). The exception to this are the $\alpha$ weights on our labels, which seem to neither improve nor impair our model's performance significantly.
Note however, that the $\alpha$ and $\beta$ weights play a strong role for the interpretability of our model (see below) and are thus useful even without influencing the raw predictive performance.}

\revision{Finally, with our labels, our model also outperforms InSight, as well as the simple logistic regression and MGP-LogReg baselines (Fig.~\ref{fig:roc_results}, bottom row). The intuition behind this result is the robustness of the model to missing data: It accounts for the data uncertainty and hence has a better performance on lower resolution and more irregular data.
Note however that on \cite{moor2019early}'s labels, the logistic regression is a stronger competitor (which was not considered in their paper), highlighting again that their prediction task is significantly easier than the one with our more realistic labels.
All these results were measuring the performance using the area under the receiver-operator characteristic curve (AUROC), but we provide additional results using the precision-recall curve (AUPRC) in the appendix (Fig.~\ref{fig:pr_results}), which qualitatively show the same observations.}

\subsubsection*{MGP interpretability}

\begin{figure*}[!h]
\includegraphics[width=\linewidth]{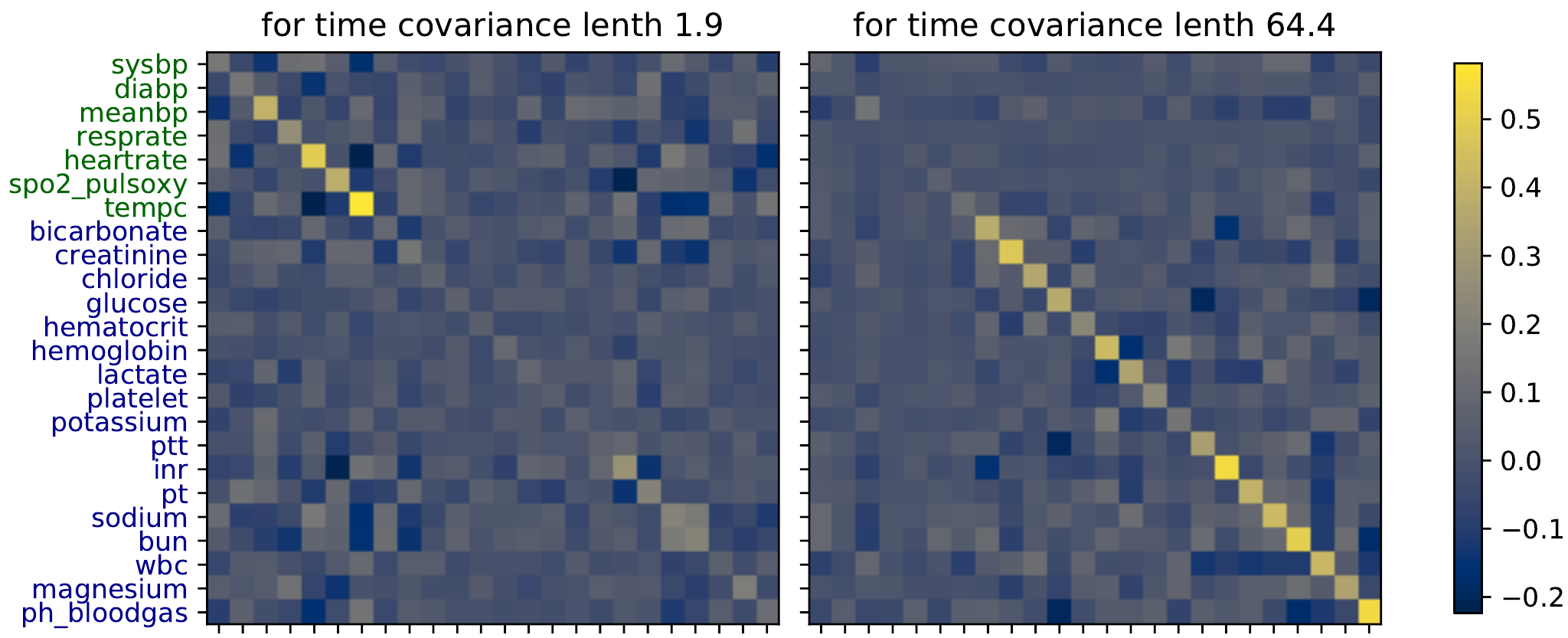}
\caption{\bf Heatmaps of the learned MGP covariance matrices between the data features for the two different smoothness clusters.}
    \label{fig:covariances}
\end{figure*}

Inspecting the learned covariances (Fig.~\ref{fig:covariances}), we notice that the two OU lengthscales converged to represent two clusters within the selected variables: a shorter lengthscale (around two hours) represents noisy data, whereas a larger lengthscale (around 64 hours) represents smoother observations. In addition, the feature covariance matrix for the short lengthscale puts more emphasis on vitals, while the one for the long lengthscale puts more emphasis on labs, fitting our initial intuition that vitals vary more rapidly. Graphically, one can observe this by inspecting the diagonals on the covariance heatmaps.

\begin{figure*}[!h]
    \centering
    \includegraphics[width=\linewidth]{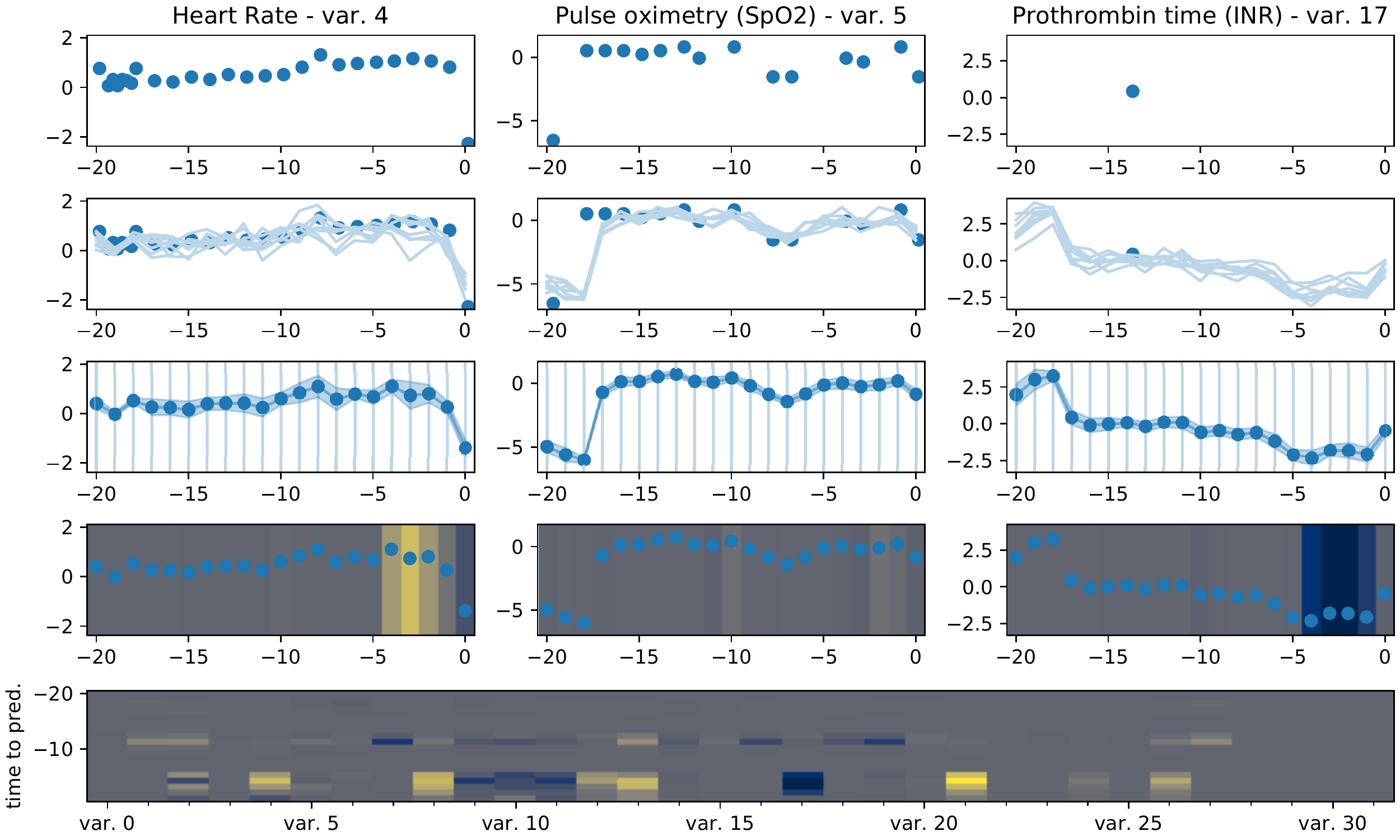}
\caption{{\bf Visualization of the journey of an exemplary patient trajectory through our proposed model architecture.} The raw features (row 1), measured at irregular time points, are interpolated by the MGP (row 2). Samples from the MGP posterior can then be aggregated into means and variances for each feature on a fixed, regularly-spaced time grid (row 3). These values are then attended to by the TCN (row 4), where positive attention weights are yellow and negative ones blue. Row 5 shows the attention weights separated by features (x-axis) and time points (y-axis).
    }
    \label{fig:model_out}
\end{figure*}

On a more granular level, the two covariance matrices also provide insights about the underlying variables.
One can for instance observe that the body temperature (\emph{tempc}) has a larger variance than the systolic and diastolic blood pressure (\emph{sysbp}, \emph{diabp}), following the general clinical intuition.
Moreover, we can observe correlations between different features, such as a negative correlation between temperature and heart rate, which also seems to coincide with the general medical expectation.
These covariances can then for instance be used by the model to extrapolate a full time series from a single INR observation with an inverse correlation to the pulse oximetry observations (Fig.~\ref{fig:model_out}).

\subsubsection*{Attention weights}

Even if our model does not outperform the MGP-TCN under all conditions, its main advantage over the baselines lies in its improved interpretability due to the attention mechanism.
Once the samples have been drawn, the weights $\boldsymbol{\alpha}$ and $\boldsymbol{\beta}$ provide us with information about the importance of different time points and features for the model's prediction.
The attention weights for an exemplary patient trajectory are depicted in Figure~\ref{fig:model_out}.

\revision{The figure shows the flow of data from a randomly chosen example patient through our model.
In the first row, we see the actual measured data. We can see that while for instance the heart rate and oximetry are measured regularly, the prothrombin time has a lot of missing values.
In the second row, we see imputations of the time series sampled from our MGP.
We see that even though the prothrombin time measurements are sparse, the MGP yields imputations with reasonably low uncertainty, thanks to information extracted by the model from the other features.
In the third row, we resample the time series from the MGP interpolations on a regular grid, which includes a mean and uncertainty estimate for each value.
Finally, in the fourth row, the TCN part of our model can assign attention weights to the different resampled measurements, which show their influence on the model's final prediction.
Positive attention weights mean that the respective feature increases the model's probability of diagnosing the patient as septic. They are shown in yellow. Negative attention weights decrease this probability and are shown in blue.
The final row of the figure shows the attention weights for all the different variables (x-axis) over time (y-axis).
}

Overall, the absolute values of $\boldsymbol{\alpha}$ are small for points further from the prediction time and increasingly larger closer to it. A good example of this behaviour is the fourth row in Figure \ref{fig:model_out}, where feature importance increases in time.
We can also see there that different features can have opposing effects on the prediction.
While the elevated heart rate close to the prediction time increases the likelihood of a sepsis prediction (first column, yellow weights), the lowered prothrombin values reduce this likelihood (third column, blue weights).
\revision{These attention weights have been deemed plausible by a clinician to whom we showed the figure, demonstrating that they can help to build trust in the model's prediction by making its decision process more relatable to trained professionals and comparable with their prior knowledge. In this particular patient, for instance, the elevated heart rate is a common symptom of sepsis \cite{morelli2015tachycardia} and thus deserves its positive attention weight, while another common symptom is an \emph{increased} prothrombin time \cite{walborn2018international}, such that the \emph{decreased} prothrombin time in this example should rightfully be regarded as speaking against the diagnosis of sepsis, as attested by its negative attention weight.}
Interestingly, the low prothrombin values are not actually measured in this example, but predicted by the MGP purely based on the other measured features and the learned covariances.

Finally, $\alpha \times \beta \times y_{\text{MC}}$ gives the individual score contribution of each feature at each time point.
These weights are shown in the last row of the figure.
It can again be seen that the attention weights are generally larger in magnitude closer to the prediction time.
Moreover, about half of the features have significant non-zero attention weights, while the others seem to not be important for the prediction in this example.

These visualizations could be used by doctors to make an informed decision about whether or not to trust the prediction of the model for each given patient, thus facilitating the interpretability and accountability that is crucial in medical applications.

\section*{Conclusion}

We have shown that current data sets for the early prediction of sepsis underestimate the true difficulty of the problem and proposed a new labelling for the MIMIC-III data set that corresponds more closely to a realistic intensive care setting.
Moreover, we have proposed a new machine learning model, the MGP-AttTCN, which outperforms the state-of-the-art approaches on the easier labels from the literature as well as on our proposed harder labels.
Additionally, our model provides an interpretable attention mechanism that will allow clinicians to make more informed decisions about trusting its predictions on a case-by-case basis.

Potential avenues for future work include a more thorough discussion with clinicians to make our proposed labels even more representative of the real-world task. Moreover, there is potential for architectural improvements, for instance by meta-learning the MGP prior \cite{fortuin2019deep, rothfuss2020pacoh}, amortizing the latent MGP inference for performance gains \cite{fortuin2019multivariate, jazbec2020scalable, ashman2020sparse, jazbec2020factorized, bing2021disentanglement}, discretizing the latent space for improved interpretability \cite{fortuin2018som, manduchi2019dpsom}, or treating the neural network parameters in a Bayesian way to improve the uncertainty estimation \cite{ciosek2019conservative, fortuin2021bayesian, garriga2021exact}.

\section*{Acknowledgments}

We would like to thank Gunnar Rätsch, Karsten Borgwardt, Michael Moor, Drago Plecko, and Nicolas Bennett for helpful discussions.


\newpage
\begin{appendices}
\onecolumn

\setcounter{table}{0}
\renewcommand{\thetable}{S\arabic{table}}

\setcounter{figure}{0}
\renewcommand{\thefigure}{S\arabic{figure}}

\def\thesection{S}

\section{\revision{Supporting information}}

 \begin{table}[!ht]
\caption{\bf \revision{Area under the ROC curve for our labels.}}
\resizebox{\linewidth}{!}{
 \begin{tabular}{c|cccccc}
 \revision{Time to onset} & \revision{MGP-Log.Reg.} & \revision{MGP-TCN} & 
 \begin{tabular}[c]{@{}c@{}}\revision{MGP-AttTCN }\\ \revision{w/o $\alpha$} \end{tabular} & 
 \begin{tabular}[c]{@{}c@{}}\revision{MGP-AttTCN } \\ \revision{w/o $\beta$}  \end{tabular} & 
 \begin{tabular}[c]{@{}c@{}}\revision{MGP-AttTCN } \\ \revision{w/ SE kernel} \end{tabular} & 
   \revision{MGP-AttTCN }\\
 \hline
 \revision{6h} & \revision{65.2 $\pm$ 1.2} & \revision{63.4 $\pm$ 1.6} & \revision{\textbf{66.7 $\pm$ 1.1}} & \revision{56.6 $\pm$ 1.1} & \revision{66.1 $\pm$ 1.4} & \revision{65.7 $\pm$ 1.6 } \\
 \revision{5h} & \revision{64.3 $\pm$ 1.3} & \revision{63.5 $\pm$ 1.2} & \revision{66.3 $\pm$ 1.4} & \revision{54.1 $\pm$ 1.0} & \revision{65.9 $\pm$ 1.2} & \revision{\textbf{66.0 $\pm$ 1.1} } \\
 \revision{4h} & \revision{66.1 $\pm$ 0.9} & \revision{63.2 $\pm$ 1.3} & \revision{66.2 $\pm$ 1.3} & \revision{56.6 $\pm$ 1.4} & \revision{64.2 $\pm$ 1.1} & \revision{\textbf{66.4 $\pm$ 1.3} } \\
 \revision{3h} & \revision{65.1 $\pm$ 0.8} & \revision{64.4 $\pm$ 1.0} & \revision{65.4 $\pm$ 0.8} & \revision{57.0 $\pm$ 1.2} & \revision{63.4 $\pm$ 1.0} & \revision{\textbf{65.7 $\pm$ 1.2} } \\
 \revision{2h} & \revision{\textbf{67.0 $\pm$ 0.7}} & \revision{65.5 $\pm$ 1.0} & \revision{66.4 $\pm$ 1.0} & \revision{57.6 $\pm$ 0.9} & \revision{63.4 $\pm$ 0.8} & \revision{66.5 $\pm$ 1.1 } \\
 \revision{1h} & \revision{\textbf{68.5 $\pm$ 0.6}} & \revision{66.7 $\pm$ 1.3} & \revision{67.9 $\pm$ 1.0} & \revision{59.1 $\pm$ 0.8} & \revision{64.9 $\pm$ 0.8} & \revision{66.6 $\pm$ 0.8 } \\
 \revision{0h} & \revision{67.5 $\pm$ 0.4} & \revision{67.3 $\pm$ 0.9} & \revision{\textbf{67.7 $\pm$ 0.7}} & \revision{62.1 $\pm$ 0.8} & \revision{65.5 $\pm$ 0.6} & \revision{65.9 $\pm$ 0.6 } \\
 \end{tabular}
 }
\label{tab:auroc}
\end{table}

\begin{table}[!ht]
\caption{\bf \revision{Area under the ROC curve for Moor et al. \cite{moor2019early} labels.}}
\resizebox{\linewidth}{!}{
 \begin{tabular}{c|cccccc}
 \revision{Time to onset} & \revision{MGP-Log.Reg.} & \revision{MGP-TCN} & 
 \begin{tabular}[c]{@{}c@{}}\revision{MGP-AttTCN }\\ \revision{w/o $\alpha$} \end{tabular} & 
 \begin{tabular}[c]{@{}c@{}}\revision{MGP-AttTCN } \\ \revision{w/o $\beta$}  \end{tabular} & 
 \begin{tabular}[c]{@{}c@{}}\revision{MGP-AttTCN } \\ \revision{w/ SE kernel} \end{tabular} & 
   \revision{MGP-AttTCN }\\
 \hline
 \revision{6h} & \revision{72.7 $\pm$ 1.4} & \revision{73.5 $\pm$ 1.2} & \revision{67.3 $\pm$ 1.9} & \revision{70.1 $\pm$ 3.1} & \revision{70.7 $\pm$ 1.4} & \revision{\textbf{76.7 $\pm$ 1.4} } \\
 \revision{5h} & \revision{73.6 $\pm$ 1.3} & \revision{74.9 $\pm$ 0.7} & \revision{67.6 $\pm$ 1.0} & \revision{69.3 $\pm$ 2.2} & \revision{70.1 $\pm$ 1.7} & \revision{\textbf{76.3 $\pm$ 1.0 }} \\
 \revision{4h} & \revision{75.2 $\pm$ 1.3} & \revision{\textbf{74.9 $\pm$ 1.0}} & \revision{68.0 $\pm$ 0.9} & \revision{72.3 $\pm$ 2.5} & \revision{70.5 $\pm$ 0.9} & \revision{74.6 $\pm$ 1.5 } \\
 \revision{3h} & \revision{76.8 $\pm$ 1.2} & \revision{76.0 $\pm$ 0.6} & \revision{69.9 $\pm$ 1.4} & \revision{72.6 $\pm$ 1.6} & \revision{72.1 $\pm$ 1.3} & \revision{76.3 $\pm$ 0.9 } \\
 \revision{2h} & \revision{79.4 $\pm$ 0.6} & \revision{\textbf{80.2 $\pm$ 0.6}} & \revision{73.6 $\pm$ 1.1} & \revision{76.3 $\pm$ 2.0} & \revision{75.2 $\pm$ 0.8} & \revision{78.9 $\pm$ 1.4 } \\
 \revision{1h} & \revision{82.6 $\pm$ 0.5} & \revision{\textbf{83.6 $\pm$ 0.4}} & \revision{77.3 $\pm$ 0.9} & \revision{78.1 $\pm$ 1.6} & \revision{80.5 $\pm$ 0.6} & \revision{82.0 $\pm$ 0.8 } \\
 \revision{0h} & \revision{83.5 $\pm$ 0.4} & \revision{\textbf{87.0 $\pm$ 0.5}} & \revision{76.9 $\pm$ 0.7} & \revision{79.3 $\pm$ 1.0} & \revision{83.6 $\pm$ 0.6} & \revision{82.3 $\pm$ 0.7 } \\
 \end{tabular}
 }
\label{tab:auroc_moor}
\end{table}

\begin{table}[!ht]
\caption{\bf \revision{Area under the precision-recall curve for our labels.}}
\resizebox{\linewidth}{!}{
 \begin{tabular}{c|cccccc}
 \revision{Time to onset} & \revision{MGP-Log.Reg.} & \revision{MGP-TCN} & 
 \begin{tabular}[c]{@{}c@{}}\revision{MGP-AttTCN }\\ \revision{w/o $\alpha$} \end{tabular} & 
 \begin{tabular}[c]{@{}c@{}}\revision{MGP-AttTCN } \\ \revision{w/o $\beta$}  \end{tabular} & 
 \begin{tabular}[c]{@{}c@{}}\revision{MGP-AttTCN } \\ \revision{w/ SE kernel} \end{tabular} & 
   \revision{MGP-AttTCN }\\
 \hline
 \revision{6h} & \revision{47.0 $\pm$ 1.5} & \revision{47.3 $\pm$ 1.9} & \revision{\textbf{50.5 $\pm$ 1.6}} & \revision{41.4 $\pm$ 1.1} & \revision{49.1 $\pm$ 1.7} & \revision{48.4 $\pm$ 1.4 } \\
 \revision{5h} & \revision{45.3 $\pm$ 1.3} & \revision{46.0 $\pm$ 1.1} & \revision{\textbf{49.6 $\pm$ 1.4}} & \revision{38.1 $\pm$ 1.3} & \revision{47.5 $\pm$ 1.1} & \revision{48.3 $\pm$ 1.9 } \\
 \revision{4h} & \revision{48.5 $\pm$ 1.1} & \revision{46.5 $\pm$ 1.6} & \revision{\textbf{49.4 $\pm$ 2.0}} & \revision{42.2 $\pm$ 1.5} & \revision{48.3 $\pm$ 1.2} & \revision{48.8 $\pm$ 1.4 } \\
 \revision{3h} & \revision{\textbf{49.1 $\pm$ 1.0}} & \revision{47.5 $\pm$ 1.0} & \revision{49.1 $\pm$ 1.1} & \revision{42.4 $\pm$ 1.2} & \revision{48.1 $\pm$ 1.1} & \revision{48.3 $\pm$ 1.2 } \\
 \revision{2h} & \revision{\textbf{50.8 $\pm$ 0.8}} & \revision{48.9 $\pm$ 1.3} & \revision{49.2 $\pm$ 1.2} & \revision{43.2 $\pm$ 0.8} & \revision{50.2 $\pm$ 1.1} & \revision{48.6 $\pm$ 0.9 } \\
 \revision{1h} & \revision{\textbf{52.3 $\pm$ 0.8}} & \revision{50.1 $\pm$ 1.5} & \revision{51.7 $\pm$ 1.2} & \revision{44.7 $\pm$ 0.9} & \revision{51.2 $\pm$ 0.8} & \revision{49.0 $\pm$ 1.0 } \\
 \revision{0h} & \revision{55.8 $\pm$ 0.6} & \revision{53.9 $\pm$ 1.2} & \revision{55.7 $\pm$ 0.8} & \revision{48.6 $\pm$ 1.1} & \revision{\textbf{57.1 $\pm$ 0.8}} & \revision{52.2 $\pm$ 0.7 } \\
 \end{tabular}
 }
\label{tab:auprc}
\end{table}

\begin{table}[!ht]
\caption{\bf \revision{Area under the precision-recall curve for Moor et al. \cite{moor2019early} labels.}}

\resizebox{\linewidth}{!}{
 \begin{tabular}{c|cccccc}
 \revision{Time to onset} & \revision{MGP-Log.Reg.} & \revision{MGP-TCN} & 
 \begin{tabular}[c]{@{}c@{}}\revision{MGP-AttTCN }\\ \revision{w/o $\alpha$} \end{tabular} & 
 \begin{tabular}[c]{@{}c@{}}\revision{MGP-AttTCN } \\ \revision{w/o $\beta$}  \end{tabular} & 
 \begin{tabular}[c]{@{}c@{}}\revision{MGP-AttTCN } \\ \revision{w/ SE kernel} \end{tabular} & 
   \revision{MGP-AttTCN }\\
 \hline
 \revision{6h} & \revision{\textbf{29.2 $\pm$ 2.0}} & \revision{20.9 $\pm$ 1.1} & \revision{23.4 $\pm$ 1.6} & \revision{22.8 $\pm$ 2.5} & \revision{24.2 $\pm$ 2.4} & \revision{24.2 $\pm$ 1.4 } \\
 \revision{5h} & \revision{\textbf{28.3 $\pm$ 1.9}} & \revision{22.3 $\pm$ 1.2} & \revision{23.9 $\pm$ 1.7} & \revision{21.4 $\pm$ 2.2} & \revision{27.0 $\pm$ 3.0} & \revision{24.2 $\pm$ 1.4  } \\
 \revision{4h} & \revision{\textbf{29.3 $\pm$ 2.0}} & \revision{22.9 $\pm$ 1.3} & \revision{23.1 $\pm$ 1.4} & \revision{22.6 $\pm$ 1.6} & \revision{26.6 $\pm$ 1.9} & \revision{25.5 $\pm$ 1.4 } \\
 \revision{3h} & \revision{\textbf{29.7 $\pm$ 1.2}} & \revision{24.5 $\pm$ 1.0} & \revision{24.5 $\pm$ 1.3} & \revision{23.6 $\pm$ 1.5} & \revision{26.5 $\pm$ 1.5} & \revision{28.6 $\pm$ 1.7 } \\
 \revision{2h} & \revision{\textbf{36.7 $\pm$ 1.8}} & \revision{31.4 $\pm$ 1.6} & \revision{29.3 $\pm$ 1.2} & \revision{28.1 $\pm$ 1.8} & \revision{33.2 $\pm$ 1.4} & \revision{31.5 $\pm$ 2.3 } \\
 \revision{1h} & \revision{\textbf{40.5 $\pm$ 0.9}} & \revision{36.8 $\pm$ 1.0} & \revision{32.5 $\pm$ 1.4} & \revision{31.6 $\pm$ 1.9} & \revision{38.2 $\pm$ 1.4} & \revision{39.4 $\pm$ 1.6 } \\
 \revision{0h} & \revision{44.4 $\pm$ 1.3} & \revision{\textbf{47.0 $\pm$ 1.3}} & \revision{33.8 $\pm$ 1.0} & \revision{36.8 $\pm$ 1.1} & \revision{41.0 $\pm$ 1.1} & \revision{43.1 $\pm$ 1.6 } \\
 \end{tabular}
 }
\label{tab:auprc_moor}
\end{table}

\begin{table}[!ht]
\caption{\bf \revision{Area under the ROC curve for our labels and the baseline dataset.}}
 \begin{tabular}{c|ccc}
  \revision{Time to onset} &  \revision{Log. Reg.} &  \revision{InSight} & \revision{MGP-AttTCN} \\
 \hline
  \revision{6h} & \revision{57.2} & \revision{54.7} & \revision{\textbf{64.05 $\pm$ 2.2}}\\
  \revision{5h} & \revision{56.6} & \revision{49.0} & \revision{\textbf{66.36 $\pm$ 1.95}}\\
  \revision{4h} & \revision{53.8} & \revision{55.9} & \revision{\textbf{67.46 $\pm$ 1.59}}\\
  \revision{3h} & \revision{53.2} & \revision{53.1} & \revision{\textbf{66.52 $\pm$ 1.65}}\\
  \revision{2h} & \revision{53.8} & \revision{55.2} & \revision{\textbf{66.99 $\pm$ 1.62}}\\
  \revision{1h} & \revision{54.3} & \revision{57.3} & \revision{\textbf{66.39 $\pm$ 1.28}}\\
  \revision{0h} & \revision{50.8} & \revision{55.9} & \revision{\textbf{64.66 $\pm$ 1.82}}\\
 \end{tabular}
\label{tab:auroc_baseline}
\end{table}

\begin{table}[!ht]
\caption{\bf \revision{Area under the ROC curve for Moor et al. \cite{moor2019early} labels and the baseline dataset.}}
 \begin{tabular}{c|ccc}
  \revision{Time to onset} &  \revision{Log. Reg.} &  \revision{InSight} & \revision{MGP-AttTCN} \\
 \hline
  \revision{6h} & \revision{\textbf{73.9}} & \revision{64.5} & \revision{72.58 $\pm$ 1.82}\\
  \revision{5h} & \revision{\textbf{73.9}} & \revision{64.4} & \revision{72.5 $\pm$ 1.59}\\
  \revision{4h} & \revision{\textbf{76.2}} & \revision{65.7} & \revision{71.78 $\pm$ 1.42}\\
  \revision{3h} & \revision{\textbf{76.5}} & \revision{74.1} & \revision{73.36 $\pm$ 1.19}\\
  \revision{2h} & \revision{\textbf{76.2}} & \revision{69.7} & \revision{75.39 $\pm$ 1.55}\\
  \revision{1h} & \revision{77.0} & \revision{74.1} & \revision{\textbf{77.71 $\pm$ 1.33}}\\
  \revision{0h} & \revision{77.3} & \revision{73.1} & \revision{\textbf{78.56 $\pm$ 1.28}}\\
  \end{tabular}
\label{tab:auroc_baseline_moor}
\end{table}

\begin{table}[!ht]
\caption{\bf \revision{Area under the Precision-Recall curve for our labels and the baseline dataset.}}
 \begin{tabular}{c|ccc}
  \revision{Time to onset} &  \revision{Log. Reg.} &  \revision{InSight} & \revision{MGP-AttTCN} \\
 \hline
  \revision{6h} & \revision{37.7} & \revision{41.4} & \revision{\textbf{41.68 $\pm$ 2.27}}\\
  \revision{5h} & \revision{37.6} & \revision{35.9} & \revision{\textbf{43.29 $\pm$ 2.5}}\\
  \revision{4h} & \revision{37.9} & \revision{45.2} & \revision{\textbf{47.7 $\pm$ 2.22}}\\
  \revision{3h} & \revision{36.2} & \revision{38.2} & \revision{\textbf{45.81 $\pm$ 2.44}}\\
  \revision{2h} & \revision{39.9} & \revision{38.8} & \revision{\textbf{46.09 $\pm$ 1.71}}\\
  \revision{1h} & \revision{42.2} & \revision{41.8} & \revision{\textbf{47.81 $\pm$ 1.75}}\\
  \revision{0h} & \revision{40.7} & \revision{46.3} & \revision{\textbf{49.31 $\pm$ 1.78}}\\
  \end{tabular}
\label{tab:auprc_baseline}
\end{table}

\begin{table}[!ht]
\caption{\bf \revision{Area under the Precision-Recall curve for Moor et al. \cite{moor2019early} labels and the baseline dataset.}}
 \begin{tabular}{c|ccc}
  \revision{Time to onset} &  \revision{Log. Reg.} &  \revision{InSight} & \revision{MGP-AttTCN} \\
 \hline
  \revision{6h} & \revision{\textbf{44.5}} & \revision{37.2} & \revision{35.52 $\pm$ 2.05}\\
  \revision{5h} & \revision{\textbf{44.3}} & \revision{37.2} & \revision{38.43 $\pm$ 2.51}\\
  \revision{4h} & \revision{\textbf{50.9}} & \revision{40.9} & \revision{38.54 $\pm$ 1.51}\\
  \revision{3h} & \revision{\textbf{55.2}} & \revision{52.2} & \revision{42.07 $\pm$ 1.22}\\
  \revision{2h} & \revision{\textbf{53.7}} & \revision{48.4} & \revision{47.08 $\pm$ 1.86}\\
  \revision{1h} & \revision{\textbf{55.8}} & \revision{52.2} & \revision{53.09 $\pm$ 1.19}\\
  \revision{0h} & \revision{55.7} & \revision{53.9} & \revision{\textbf{57.87 $\pm$ 1.86}}\\
  \end{tabular}
\label{tab:auprc_baseline_moor}
\end{table}

\begin{figure*}[!h]
    \centering
	\includegraphics[width=\linewidth]{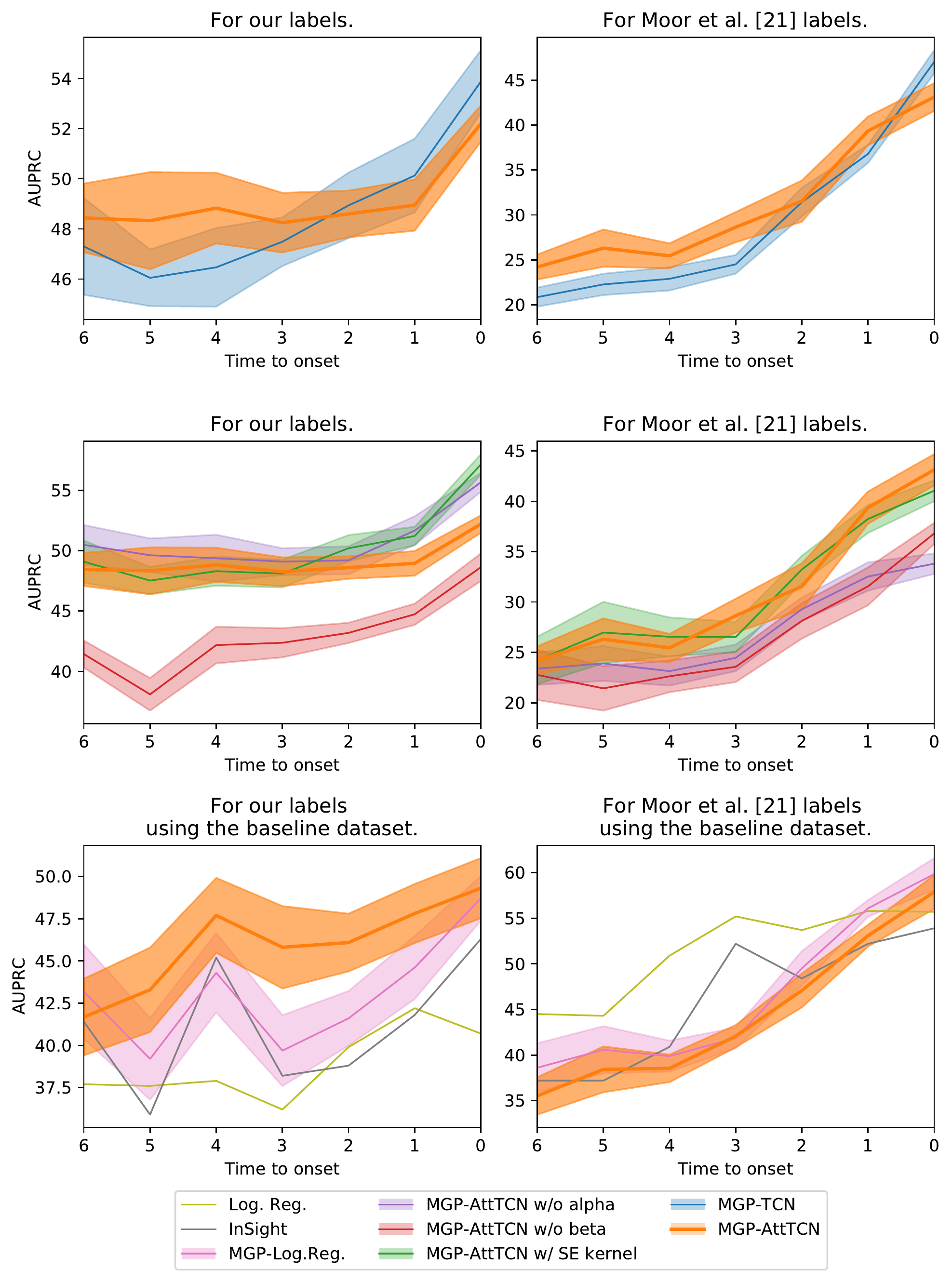}
    \caption{{\bf \revision{Area under the Precision-Recall curve of different models.}}}
    \label{fig:pr_results}
\end{figure*} \end{appendices}

\end{document}